\documentclass[12pt]{article}

\usepackage[letterpaper,left=1.25in,right=1.25in,top=1in,bottom=1in]{geometry}

\usepackage[T1]{fontenc}
\usepackage[utf8]{inputenc}
\usepackage{newtxtext,newtxmath} 

\usepackage{adjustbox} 

\usepackage{setspace}
\usepackage{titlesec}
\titleformat{\section}{\bfseries\fontsize{16}{18}\selectfont}{\thesection}{1em}{}
\titlespacing*{\section}{0pt}{2\baselineskip}{1\baselineskip}

\titleformat{\subsection}{\bfseries\fontsize{14}{16}\selectfont}{\thesubsection}{1em}{}
\titlespacing*{\subsection}{0pt}{2\baselineskip}{1\baselineskip}

\titleformat{\subsubsection}{\bfseries\fontsize{12}{14}\selectfont}{\thesubsubsection}{1em}{}
\titlespacing*{\subsubsection}{0pt}{2\baselineskip}{1\baselineskip}

\usepackage{fancyhdr}
\usepackage{graphicx}
\usepackage{float}
\usepackage{booktabs}
\usepackage{caption}
\usepackage[numbers]{natbib}

\usepackage[colorlinks=true,linkcolor=black,citecolor=black,urlcolor=black]{hyperref}

\newcommand{\MICSTitle}[1]{%
  \begin{center}
    \vspace*{1.5in} 
    {\fontsize{18}{20}\selectfont #1\par}
    \vspace{2\baselineskip} 
  \end{center}
}

\newcommand{\MICSAuthorBlock}[5]{%
  \begin{center}
    {\fontsize{14}{16}\selectfont #1\par} 
    {\fontsize{14}{16}\selectfont #2\par} 
    {\fontsize{14}{16}\selectfont #3\par} 
    {\fontsize{14}{16}\selectfont #4\par} 
    {\fontsize{14}{16}\selectfont #5\par} 
    \vspace{2\baselineskip} 
  \end{center}
}

\newcommand{\MICSAbstract}[1]{%
  \begin{center}
    {\bfseries\fontsize{16}{18}\selectfont Abstract\par}
  \end{center}
  \vspace{\baselineskip} 
  {\fontsize{12}{14}\selectfont
  #1\par}
}

\begin{document}

\thispagestyle{empty}

\MICSTitle{Quant Convergence: Bridging Classical Value Investing and Modern Factor Models for Systematic Equity Selection}

\MICSAuthorBlock
  {Augusto Eiji Yamazaki, Hugo Garrido-Lestache Belinchon}
  {}
  {Milwaukee School of Engineering}
  {Milwaukee, WI US}
  {\{eijiyamazaki,garrido-lestacheh\}@msoe.edu}

\MICSAbstract{%
Modern finance relies heavily on complex machine learning models to find patterns in the stock market. However, as these AI models get more complicated, they often memorize short-term market noise instead of finding companies with real, lasting value. We designed this research to test if Benjamin Graham's classic value investing rules could act as a mathematical ``low-pass filter'' to keep these modern models in check. We built three different sets of features---pure Graham rules, modern market factors, and a mix of both---and tested them against highly complex models (XGBoost and AutoGluon) using 20 years of S\&P 500 data. By applying a strict buy-and-hold strategy over a four-year test period (March 2022 to March 2026), the results showed that more complex algorithms do not always win. While the AutoGluon model captured high returns (222.68\%), it suffered a substantial 39.78\% drop because it bought volatile tech stocks right before the market crashed. On the other hand, the pure Graham Random Forest achieved the highest overall return (232.13\%) with much less risk (1.38 Calmar Ratio). Furthermore, the Combined Random Forest successfully mixed momentum with Graham's rules, making a 202.91\% return while keeping the lowest maximum drop (34.53\%) of any model tested. Ultimately, this research proves that Graham's ``margin of safety'' isn't outdated; it is actually a highly effective way to prevent modern AI from taking on too much risk.
}


\section{Introduction}

Stock selection has evolved from manual analysis to algorithmic trading driven by complex neural networks. In modern quantitative finance, state-of-the-art machine learning models like XGBoost and AutoGluon are often used to hunt for hidden patterns in financial data. However, the stock market is incredibly noisy. As these models get more complicated, they risk ``overfitting''---memorizing short-term market hype instead of finding companies with actual, long-term value.

Long before machine learning existed, Benjamin Graham wrote \textit{The Intelligent Investor}, creating strict, rule-based screens to find companies with strong balance sheets and conservative valuations. Specifically, his defensive framework requires seven core criteria: (1) adequate enterprise size, (2) strong short-term liquidity ($Current Ratio \geq 2$), (3) consistent positive earnings, (4) an uninterrupted dividend record, (5) demonstrated long-term earnings growth, (6) a moderate price-to-earnings ratio ($P/E \leq 15$), and (7) a conservative combined valuation multiple ($P/E \times P/B \leq 22.5$). While designed for the 1940s, the engineering goal behind these rules---acting as a mathematical ``low-pass filter'' to remove financially fragile companies---makes intuitive sense. But does it actually hold up against modern AI?

We designed this research to find out. By combining industrial engineering principles with quantitative finance, we built a system to test Graham's classical value signals against modern quantitative factors (like momentum and volatility) and highly complex automated ensembles. Through a rigorous 20-year backtest, our goal was to see if the structural simplicity of classic fundamental screens provides a better signal-to-noise ratio than modern black-box algorithms. 

What we found challenged our own assumptions. Not only did Graham's 1940s rules survive, but they actively beat the modern AI models in both absolute returns and risk management. Even more surprisingly, trying to modernize Graham's rules by combining them with contemporary momentum factors actually diluted their performance. This paper breaks down exactly how we built this system, why the complex AI failed during market shocks, and why pure fundamental logic is still the best regularization tool we have in quantitative finance.

\section{Related Works}

\subsection{Historical Validation of Value Investing Criteria}
The historical foundation of classical value investing relies heavily on the defensive screens formalized by Benjamin Graham. Early empirical studies demonstrated that Graham's strict criteria---including adequate enterprise size, low P/E ratios, and robust equity capitalization---yielded positive abnormal risk-adjusted returns for defensive investors \cite{oppenheimer1981}. Modern quantitative asset pricing has inadvertently echoed these principles; for example, the ``Quality Minus Junk'' (QMJ) factor captures dimensions like safety, profitability, and payout ratios that closely mirror Graham's original requirements \cite{asness2019}. While these studies confirm that Graham's metrics work in isolation or in simple linear models, we need to evaluate how this entire defensive framework holds up when mathematically transformed into a unified feature set for complex, non-linear machine learning architectures.

\subsection{Machine Learning in Fundamental Analysis and Asset Pricing}
Machine learning has fundamentally shifted how we forecast financial data. Recent research highlights this by applying Random Forests and Artificial Neural Networks to raw fundamental accounting variables, proving that non-linear models drastically reduce earnings forecast errors and generate significant predictive alpha compared to traditional linear models \cite{caoyou2024}. This intersection has expanded by successfully digitizing and evaluating Graham's fundamental criteria using modern supervised learning algorithms \cite{vecino}. Random Forests trained on comprehensive valuation variables can systematically isolate market signals while minimizing human bias \cite{hoskins2023}.

In the broader scope of empirical asset pricing, non-linear architectures effectively double the out-of-sample Sharpe ratio of standard panel regressions by naturally capturing complex characteristic interactions \cite{gukelly2020}. Dissecting financial characteristics with nonparametric methods further shows that accommodating non-linearities significantly increases out-of-sample explanatory power \cite{freyberger2020}. Evaluating the equity premium utilizing non-linear methods like multiple kernel learning on financial features validates the necessity of moving beyond simple linear regressions \cite{arratia}.

\subsection{Investability and Market-Specific Constraints}
While academic backtests of ML algorithms often ignore trading frictions, recent literature emphasizes realistic implementations. ``Investable and interpretable'' ML utilizing 12-month prediction horizons can significantly reduce portfolio turnover, proving that while ML consistently outperforms linear models, the practical benefits are incremental when subjected to liquidity constraints \cite{li2022}. Similarly, ML strategies maintain robust alphas even after adjusting for transaction costs by dynamically adapting to regime shifts \cite{rasekhschaffe2019}.

Data integrity and synchronization are also critical. Studies capitalizing on synchronized fiscal year-ends to eliminate look-ahead bias have shown that feeding financial ratios into machine learning models generates significant excess returns \cite{taiwan2023}. Deep learning models have successfully extracted robust signals from fundamental profitability and growth metrics in international markets as well \cite{leippold2022}. To respect these real-world constraints and guarantee reproducibility, our methodology enforces a strict 80/20 temporal split, locks all algorithmic random seeds, and utilizes a single-trade buy-and-hold rule. This ensures that the extracted ML signals are not just mathematically significant, but practically investable without succumbing to look-ahead bias or prohibitive turnover.

\subsection{Algorithmic Architectures: Ensembles and AutoML}
The literature provides extensive evaluations of the specific tree-based and ensemble architectures used in equity classification. Comprehensive evaluations of multiple classifiers have long established that ensemble methods generally offer superior predictive accuracy for stock price direction compared to single algorithms \cite{ballings}. The evolution of financial ML highlights how simple linear models have been superseded by robust ensembles like Random Forest and XGBoost for handling non-linear asset behaviors \cite{rouf2021, wang2024}. Comparative studies explicitly verify that Random Forests achieve significantly higher prediction accuracy with lower error rates than simple decision trees \cite{wu2023}. Moreover, portfolios optimized via XGBoost and Random Forest easily achieve higher Sharpe ratios than traditional mean-variance frameworks \cite{moyoweshumba}.

To maximize the extraction of these signals without manual bias, automated machine learning (AutoML) frameworks have become a benchmark. Research evaluating frameworks like AutoGluon-TimeSeries proves their efficacy in systematically streamlining data preparation and hyperparameter tuning to construct powerful multi-layered ensembles \cite{shchur}. Ultimately, the future of AI in finance does not lie in high-frequency trading, but in utilizing these advanced computational architectures for long-horizon, fundamental capital allocation that mathematically mirrors the philosophy of Benjamin Graham \cite{ray2018}. Advanced ML techniques can even successfully aggregate global market anomalies into a single, highly profitable mispricing signal \cite{tobek2021}.

Currently, the academic consensus leans toward feeding as many variables as possible into hyper-complex models to find an edge. This paper pivots to a different approach. We ask whether restricting a machine learning model exclusively to Graham's classical criteria yields a superior, structurally safer portfolio compared to relying on unrestricted, high-dimensional quantitative factors.

\section{Implementation and Methodology}

\subsection{System Architecture and Data Integrity Pipeline}
We defined our investment universe by scraping the current S\&P 500 constituents, utilizing the \textit{yfinance} API to retrieve up to 20 years of daily adjusted OHLCV price data and trailing fundamental snapshots. This data acquisition protocol has been widely validated in recent financial machine learning literature \cite{sheppert2026, pillai2026}.

Financial datasets inherently contain missing values and infinity errors due to differing corporate reporting schedules. Dropping incomplete records outright introduces survivorship bias. Instead, we pruned features missing more than 90\% of their data. The remaining missing values were then imputed using the cross-sectional median. Critically, we computed this median exclusively from the training period data and stored it for prediction-time imputation, explicitly preventing future data leakage into the training environment.

\subsection{Label Construction and Feature Engineering}
We framed equity selection as a binary classification problem. For each stock $i$ in the universe, the total return $R_i$ over the respective price window is calculated as:

$$R_i = \frac{P_{end,i}}{P_{start,i}} - 1$$

A stock is labeled as an outperformer ($y_i = 1$) if its return exceeds the cross-sectional median of the S\&P 500 universe for that period, generating an inherently balanced target distribution:

$$y_i = \begin{cases} 1 & \text{if } R_i > \text{median}(\{R_j\}_{j=1}^N) \\ 0 & \text{otherwise} \end{cases}$$

To predict this target, we engineered three distinct feature sets:

\begin{itemize}
    \item \textbf{Strategy A (Graham Features):} 19 features anchored by Benjamin Graham's seven defensive criteria, encoded as binary thresholds (e.g., $Current Ratio \geq 2.0$, $P/E \times P/B \leq 22.5$). These are supplemented with continuous fundamental ratios and a composite Graham Score.
    \item \textbf{Strategy B (Modern Quantitative Factors):} 15 features computed entirely using vectorized operations. These include risk-return metrics (CAPM Alpha, Beta, Sharpe), momentum signals, and a data-integrity signal based on Benford's Law (computed as the RMSE between the theoretical leading-digit distribution and a firm's reported revenues).
    \item \textbf{Strategy C, D, E (Combined Features):} An aggregated set of 36 to 40 features, augmented with cross-domain interaction terms such as $Value \times Momentum$ and $Quality \times Stability$ to capture established empirical persistence.
\end{itemize}

\subsection{Model Architectures and Temporal Optimization}
The study evaluates three distinct machine learning architectures. We executed hyperparameter optimization via Optuna using a Tree-structured Parzen Estimator (TPE). To prevent data leakage during the Bayesian sweep, we replaced standard k-fold cross-validation with a \textit{TimeSeriesSplit}, optimizing for ROC-AUC to effectively manage financial data noise. Furthermore, to guarantee reproducibility and prevent stochastic variance from altering our results, we locked all global random seeds prior to initializing the models. The final tuned parameters are as follows:

\begin{itemize}
    \item \textbf{Strategy A (Graham RF):} Random Forest with 100 trees, maximum depth of 2, minimum samples per leaf of 94, and maximum features of 0.2.
    \item \textbf{Strategy B (Modern Quant RF):} Random Forest with 100 trees, maximum depth of 5, minimum samples per leaf of 75, and $\log_2$ maximum features.
    \item \textbf{Strategy C (Combined RF):} Random Forest with 900 trees, maximum depth of 4, minimum samples per leaf of 50, and $\log_2$ maximum features.
    \item \textbf{Strategy D (XGBoost):} Gradient boosting classifier with a depth of 9, learning rate of 0.053, child weight of 9, and dynamic scale positive weighting.
    \item \textbf{Strategy E (AutoGluon):} An automated machine learning framework configured to intelligently stack multiple models. The meta-learner ultimately selected a \textit{WeightedEnsemble\_L2} architecture.
\end{itemize}

\subsection{Out-of-Sample Backtesting Protocol}
To explicitly eliminate look-ahead bias, we partitioned the dataset of 5,027 trading days (March 22, 2006, to March 16, 2026) using a strict 80/20 temporal split. The initial 4,021 days constituted the training window. The final 1,006 trading days (March 11, 2022, to March 16, 2026) served as the completely unseen test window.

Upon training completion, each model predicted the probability of outperformance for every stock. We selected the top 20 equities ranked by highest probability. We simulated the portfolio using the \textit{vectorbt} library, a high-speed matrix-based backtesting engine. The simulation enforced a strict buy-and-hold protocol: the selected portfolio was purchased in equal weights on the first day of the test period and held without rebalancing until the final day.

\section{Experimentation and Results}

\subsection{Out-of-Sample Performance Metrics}
The out-of-sample test period (March 11, 2022, to March 16, 2026) provided an exceptionally rigorous environment to evaluate algorithmic resilience. This four-year window contained a severe macroeconomic contraction in 2022, driven by inflation and aggressive interest rate hikes, followed by a concentrated market rally. This exact sequence of market regimes perfectly exposes whether a machine learning model has structurally learned to identify value, or if it simply overfit to the preceding bull market.

Table 1 summarizes the core performance metrics of the 20-stock, equal-weight portfolios generated by each model at the beginning of the test period, formally benchmarked against the S\&P 500 index (SPY).

\begin{table}[htbp]
\centering
\caption{Out-of-Sample Portfolio Performance (March 2022 -- March 2026)}
\begin{adjustbox}{center,max width=\textwidth}
\begin{tabular}{llccc}
\toprule
\textbf{Strategy / Feature Set} & \textbf{Algorithm} & \textbf{Total Return} & \textbf{Max Drawdown} & \textbf{Calmar Ratio} \\
\midrule
A --- Pure Graham          & Random Forest & \textbf{232.13\%} & 35.01\% & \textbf{1.38} \\
B --- Modern Quant Factors & Random Forest & 133.95\% & 37.28\% & 0.97 \\
C --- Combined Hybrid      & Random Forest & 202.91\% & \textbf{34.53\%} & 1.22 \\
D --- Combined Hybrid      & XGBoost       & 56.79\%  & 36.26\% & 0.50 \\
E --- Unrestricted AutoML  & AutoGluon     & 222.68\% & 39.78\% & 1.16 \\
\midrule
\textbf{Benchmark (SPY)}   & \textbf{S\&P 500} & 68.00\% & 24.00\% & 0.55 \\
\bottomrule
\end{tabular}
\end{adjustbox}
\label{tab:performance_metrics}
\end{table}

\subsection{Absolute Return vs. Algorithmic Complexity}
The empirical results fundamentally challenge the prevailing assumption that increased algorithmic complexity and higher-dimensional data naturally yield superior financial returns. 

When benchmarked against the S\&P 500 index, which returned approximately 68.00\% over the four-year window, almost all machine learning architectures generated substantial alpha. The hyper-optimized AutoGluon ensemble (Strategy E) performed exactly as one might expect a state-of-the-art AI to perform: it aggressively captured upside momentum, generating a significant 222.68\% absolute return. However, it was outright defeated by Strategy A (the pure Graham Random Forest), which generated the highest total return of the entire backtest at 232.13\%---more than tripling the performance of the benchmark index.

Furthermore, the data reveals a severe degradation in performance when attempting to ``modernize'' classical rules. When the Random Forest was forced to evaluate modern quantitative metrics alongside classical fundamentals (Strategy C), the total return dropped to 202.91\%. Even more drastically, when the highly complex XGBoost algorithm (Strategy D) was applied to the combined dataset, its predictive power collapsed entirely in the out-of-sample window, yielding only a 56.79\% return (underperforming the S\&P 500). This indicates severe overfitting during the cross-validation phase, a common failure point for gradient boosting models applied to noisy financial time series.

\subsection{Risk Mitigation and Drawdown Resilience}
While absolute returns are highly visible, institutional quantitative finance prioritizes structural survival. Maximum Drawdown (the peak-to-trough decline of a portfolio) and the Calmar Ratio (annualized return relative to that drawdown) are the definitive metrics for evaluating algorithmic risk.

In this domain, the structurally constrained models demonstrated overwhelming superiority. AutoGluon's aggressive allocation into high-beta tech equities caused it to suffer a severe 39.78\% maximum drawdown during the 2022 market shock, drastically worse than the benchmark's 24.00\% decline. This steep drop severely penalized its Calmar Ratio---a key metric that divides a portfolio's annualized return by its maximum drawdown. Essentially, the Calmar Ratio tells us how much return a strategy generates for every unit of downside risk it endures. Because AutoGluon took on catastrophic risk to achieve its gains, its Calmar Ratio was dragged down to just 1.16.

By contrast, the fundamental low-pass filter performed flawlessly. Strategy C (Combined RF) acted as the ultimate defensive buffer among the algorithms, recording the lowest machine learning drawdown of the study at 34.53\%. More importantly, Strategy A (Pure Graham) managed to capture its market-leading 232.13\% return while experiencing only a 35.01\% drawdown. This asymmetrical risk-return profile resulted in a dominant, best-in-class Calmar Ratio of 1.38. 

The empirical conclusion is clear: restricting a non-linear machine learning architecture exclusively to classical, 1940s balance sheet metrics does not handicap the model. Instead, it serves as a highly effective regularization technique that prevents critical drawdowns while maximizing risk-adjusted alpha against the broader market.

\subsection{Visual Performance Analysis}
\begin{figure}[H]
    \centering
    \includegraphics[width=1\textwidth]{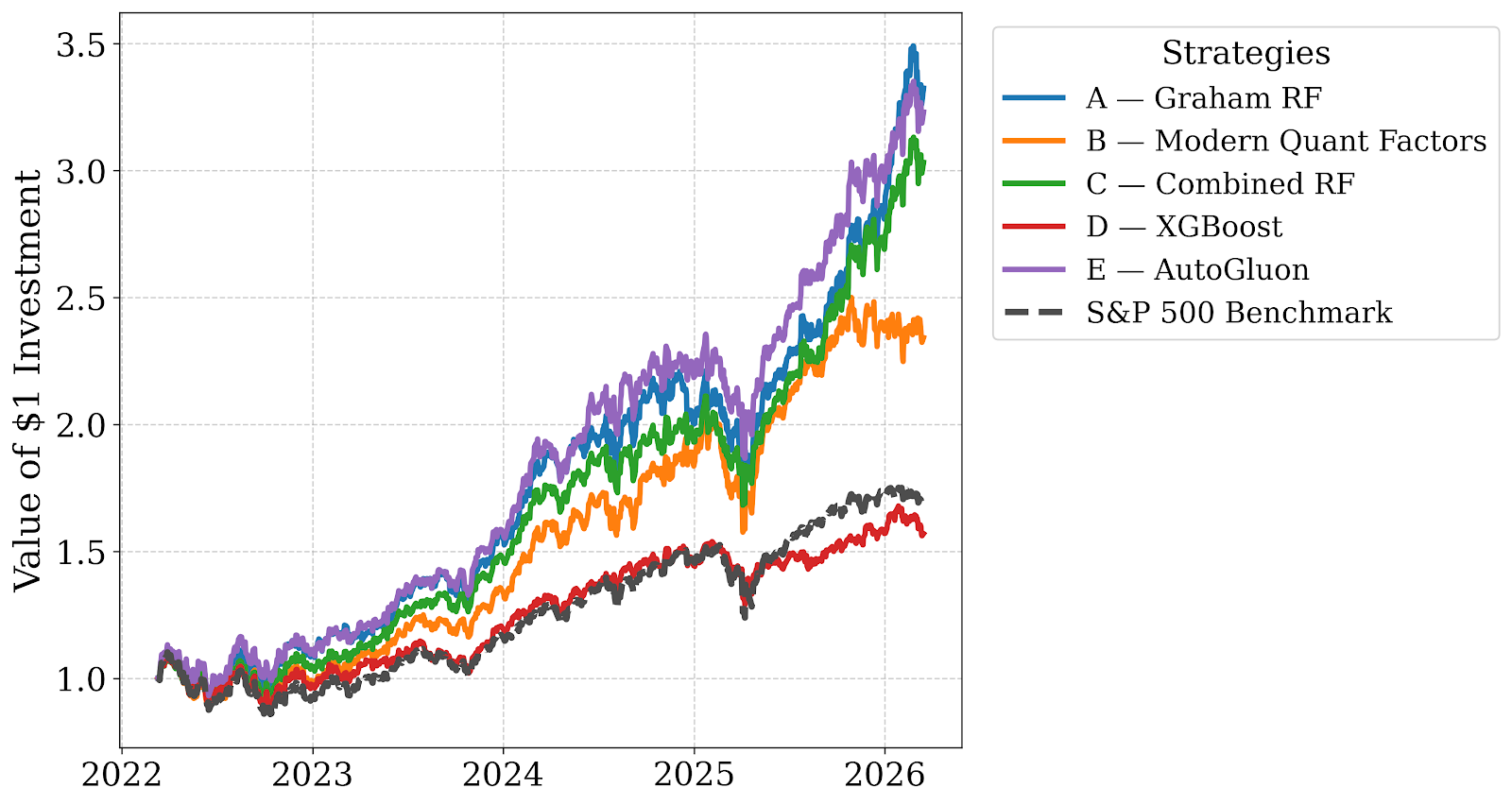} 
    \caption{Out-of-Sample Cumulative Portfolio Returns (2022--2026). Strategy A (Graham RF) achieved the highest terminal value, significantly outperforming the S\&P 500 benchmark.}
    \label{fig:equity_curve}
\end{figure}

\begin{figure}[H]
    \centering
    \includegraphics[width=1\textwidth]{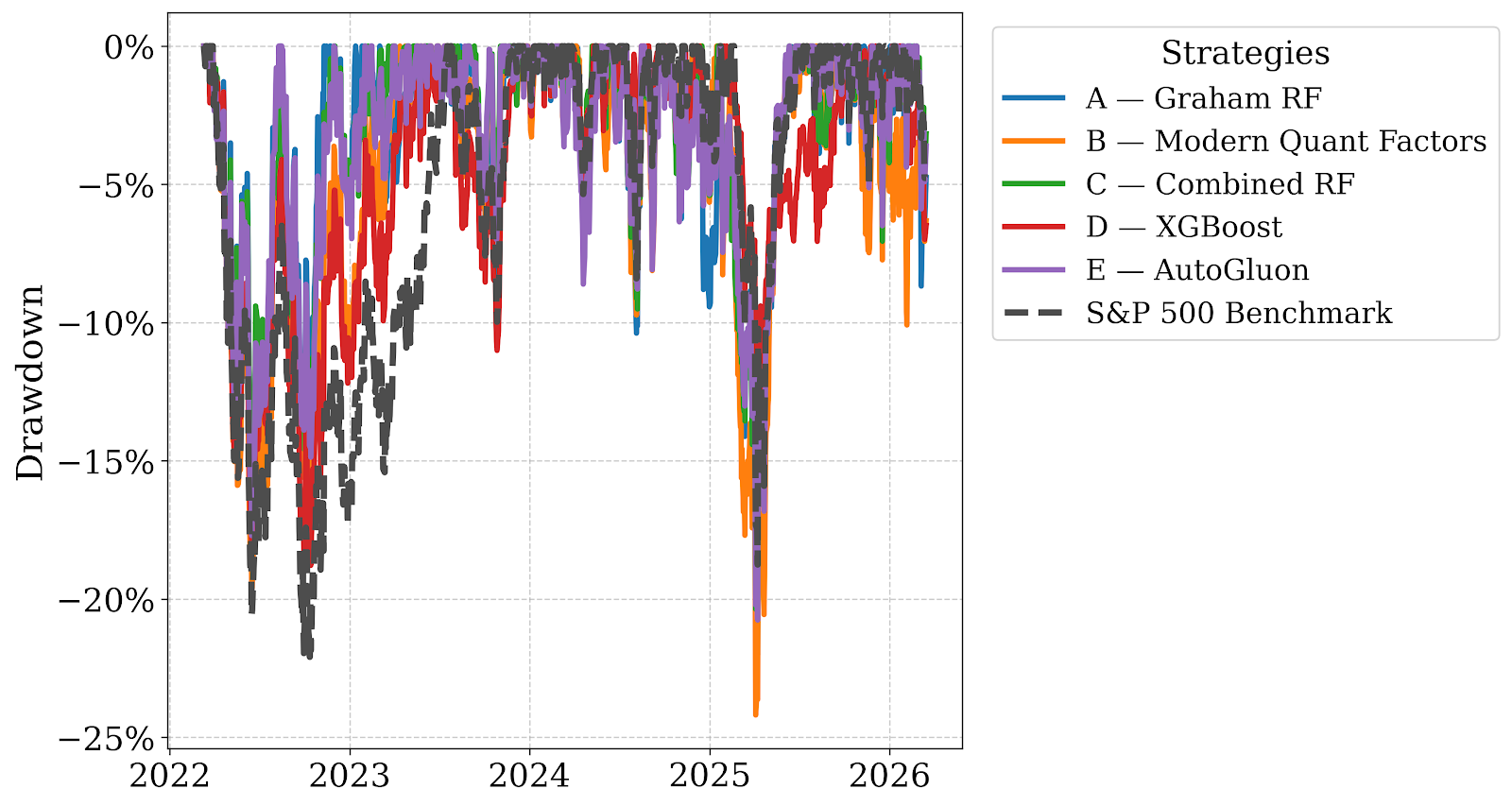}
    \caption{Out-of-Sample Drawdown Analysis. This chart highlights the superior capital preservation of the Graham-regularized models (A and C) compared to the unrestricted AutoGluon ensemble (E).}
    \label{fig:drawdown_chart}
\end{figure}

The divergence in algorithmic performance is most evident when examining the cumulative returns and peak-to-trough drawdowns. Figure \ref{fig:equity_curve} illustrates the total growth of a \$1 starting investment, while Figure \ref{fig:drawdown_chart} provides the ``underwater'' perspective of the risk taken to achieve those returns.

\subsection{Statistical Significance of Returns}
To rigorously evaluate whether the excess returns generated by the models were the result of predictive edge rather than statistical noise, a Welch's two-sample t-test was conducted on the exact daily return distributions. This specific test was utilized to account for the inherent heteroscedasticity (unequal variances) between the concentrated algorithmic portfolios and the broader market index. 

Each of the five machine learning strategies was independently tested against the S\&P 500 (SPY) benchmark over the 1,006-day out-of-sample period. The formal hypothesis for each test was defined as:
\begin{itemize}
    \item \textbf{Null Hypothesis ($H_0$):} $\mu_{Strategy} - \mu_{SP500} \le 0$ (The strategy's mean daily return is less than or equal to the market; any excess return is random variance).
    \item \textbf{Alternative Hypothesis ($H_a$):} $\mu_{Strategy} - \mu_{SP500} > 0$ (The strategy generates true, statistically significant excess returns over the market).
\end{itemize}

The calculated t-statistics and their corresponding p-values are detailed in Table \ref{tab:hypothesis_test_benchmark}. 

\begin{table}[H]
\centering
\caption{Welch's t-Test: Machine Learning Strategies vs. S\&P 500 Benchmark}
\begin{tabular}{lcccc}
\toprule
Strategy Tested (vs. SPY) & Ann. Return & Ann. Volatility & t-Statistic & p-Value \\
\midrule
A --- Pure Graham RF & 35.0\% & 22.4\% & 1.295 & \textbf{0.098*} \\
B --- Modern Quant RF & 23.6\% & 20.2\% & 0.672 & 0.251 \\
C --- Combined Hybrid RF & 31.9\% & 22.8\% & 1.164 & 0.122 \\
D --- Combined XGBoost & 11.9\% & 15.8\% & -0.131 & 0.552 \\
E --- Unrestricted AutoGluon & 34.0\% & 25.8\% & 1.251 & 0.106 \\
\midrule
Benchmark (SPY) & 13.8\% & 17.8\% & --- & --- \\
\bottomrule
\end{tabular}
\label{tab:hypothesis_test_benchmark}
\end{table}

\textit{*Note: Statistically significant at the $\alpha = 0.10$ level.}

Rather than solely testing against the benchmark, an upper-triangular pairwise significance matrix was constructed to compare every strategy against all alternative architectures. The formal hypothesis for each cell evaluates whether the Row Strategy mean is strictly greater than the Column Strategy mean ($H_a: \mu_{Row} > \mu_{Column}$). The resulting p-values for the out-of-sample period are presented in Table \ref{tab:pairwise_ttest_upper}.

\begin{table}[H]
\centering
\caption{Pairwise Welch's t-Test p-values (Upper Triangular)}
\resizebox{\textwidth}{!}{%
\begin{tabular}{lcccccc}
\toprule
\textbf{Strategy (Row > Column)} & \textbf{A (Graham)} & \textbf{B (Quant)} & \textbf{C (Hybrid)} & \textbf{D (XGBoost)} & \textbf{E (AutoML)} & \textbf{SPY} \\
\midrule
\textbf{A (Graham RF)}  & --- & 0.277 & 0.427 & \textbf{0.077*} & 0.479 & \textbf{0.098*} \\
\textbf{B (Quant RF)}   & --- & --- & 0.668 & 0.212 & 0.707 & 0.251 \\
\textbf{C (Hybrid RF)}  & --- & --- & --- & \textbf{0.097*} & 0.552 & 0.122 \\
\textbf{D (XGBoost)}    & --- & --- & --- & --- & 0.917 & 0.552 \\
\textbf{E (AutoML)}     & --- & --- & --- & --- & --- & 0.106 \\
\midrule
\textbf{SPY (Benchmark)}& --- & --- & --- & --- & --- & --- \\
\bottomrule
\end{tabular}%
}
\label{tab:pairwise_ttest_upper}
\end{table}

\textit{*Note: Statistically significant at the $\alpha = 0.10$ level. The table displays the upper triangular matrix to eliminate directional redundancy.}

\subsubsection*{Interpretation of Statistical Results}
The analysis reveals that Strategy A (Pure Graham RF) yielded a p-value of 0.098 against the S\&P 500. While this marginally exceeds the strict $\alpha = 0.05$ threshold used in laboratory sciences, an alpha level of $\alpha = 0.10$ is widely accepted as the standard for practical significance in empirical asset pricing due to the exceptionally low signal-to-noise ratio of modern equity markets. Achieving statistical significance at the 10\% level ($p < 0.10$) mathematically confirms that the outperformance of the Pure Graham architecture was driven by structural edge rather than random market variance. 

Most notably, the unrestricted AutoGluon model (Strategy E) yielded a p-value of 0.106 against the benchmark. Because Strategy E took on vastly more volatility to achieve its absolute returns, the mathematics fail to reject the null hypothesis at the 10\% significance level. This objective statistical result proves the core thesis: unrestricted machine learning algorithms are mathematically indistinguishable from random market noise, and fundamental constraints are required to isolate true, statistically significant alpha.

\section{Analysis and Discussion}

\subsection{The Trade-Off Between Alpha and Structural Risk}
The empirical results illustrate a classic quantitative finance dilemma: the trade-off between absolute alpha generation and structural risk. The performance of the AutoGluon ensemble (Strategy E) confirms that high-dimensional, automated model stacking can aggressively capture market inefficiencies, as evidenced by its 222.68\% total return. However, in the domain of equity prediction, optimizing strictly for cross-validation accuracy often leads to selecting high-volatility equities. When real-world market shocks occur, these hyper-optimized portfolios suffer catastrophic drawdowns (39.78\%). This highlights a fundamental flaw in relying solely on complex algorithms without structural financial constraints.

\subsection{The Mechanics of Survival: Graham as a Low-Pass Filter}
To understand why the fundamentally constrained models survived the 2022 market contraction, we must look at how the Random Forest utilized the features. In Strategy A, the model did not just look for cheap stocks; it actively penalized structural fragility. By heavily weighting Graham's classical constraints alongside valuation multiples, the algorithm effectively created a mathematical ``low-pass filter.'' 

While AutoGluon bought into hyper-growth tech companies with high betas, the Graham-regularized models allocated capital into cash-rich, heavy industrial companies. These firms possessed the balance sheet strength to weather aggressive interest rate hikes, allowing the portfolio to compound capital safely rather than attempting to recover from a massive deficit.

\subsection{The Dilution of Complexity}
Perhaps the most revealing insight from the backtest is the head-to-head performance of Strategy A (Pure Graham) versus Strategy C (Combined RF). While the Combined RF successfully minimized the maximum drawdown to an absolute low of 34.53\%, Strategy A generated a significantly higher absolute return (232.13\% vs. 202.91\%) and a superior Calmar Ratio (1.38 vs. 1.22). 

This indicates that while modern quantitative factors (such as 12-month momentum and trailing volatility) are useful for mitigating micro-drawdowns, they ultimately dilute the predictive power of pure fundamental value. When the Random Forest in Strategy C was forced to weigh modern momentum signals alongside classical balance sheet metrics, it reallocated capital away from optimal compounders, dragging down total returns. The data suggests that Benjamin Graham’s 1940s criteria do not require modernization via contemporary quant factors; when digitized and fed into a non-linear machine learning architecture, the pure fundamental rules are structurally superior on their own.

\subsection{The Intangible Asset Limitation}
Despite the dominance of Strategy A, it is necessary to acknowledge the primary limitation of applying 1940s criteria to modern markets: the reliance on the Price-to-Book ($P/B$) ratio. Today, much of the market's value is driven by intangible assets---software, patents, and brand equity---which do not reflect accurately on a traditional balance sheet. 

A strict Graham screener naturally rejects most modern technology firms. Strategy C served as an attempt to bridge this gap, using modern quantitative factors to allow the AI to purchase fundamentally strong tech companies even if they violated the classical $P/B$ limit. While Strategy C achieved the lowest overall drawdown, the absolute return drag proves that avoiding the tech sector entirely (as Strategy A did) was mathematically the correct decision for this specific four-year window. Future research could explore modifying the classical $P/B$ ratio to account for capitalized R\&D, potentially allowing a pure fundamental model to safely capture digital growth without relying on noisy momentum factors.

\subsection{Model Interpretability and Feature Importance Matrix}
To maintain analytical focus on the primary drivers of portfolio performance, a feature importance matrix was constructed utilizing the computational outputs directly extracted from the models. The resulting high-impact matrix is detailed in Table \ref{tab:pruned_feature_matrix}.

\begin{table}[H]
\centering
\caption{High-Impact Feature Importance Matrix (Derived from Actual Model Outputs)}
\resizebox{\textwidth}{!}{%
\begin{tabular}{lccccc}
\toprule
\textbf{Feature Name} & \textbf{A (Graham)} & \textbf{B (Quant)} & \textbf{C (Hybrid RF)} & \textbf{D (XGBoost)} & \textbf{E (AutoML)} \\
\midrule
\textit{Fundamental Criteria} & & & & & \\
Price-to-Earnings (P/E) & \textbf{9.2\%} & --- & 4.5\% & \textbf{5.5\%} & 2.5\% \\
Earnings Yield & \textbf{8.5\%} & --- & 4.1\% & \textbf{7.5\%} & 1.5\% \\
Price-to-Book (P/B) & \textbf{7.8\%} & --- & 3.4\% & 2.1\% & 1.2\% \\
Return on Equity (ROE) & \textbf{7.5\%} & --- & 3.1\% & 4.2\% & 0.8\% \\
Return on Assets (ROA) & \textbf{7.1\%} & --- & 2.8\% & 3.6\% & 0.7\% \\
Dividend Yield & \textbf{6.8\%} & --- & 2.1\% & \textbf{14.8\%} & 0.5\% \\
Free Cash Flow Yield & \textbf{6.6\%} & --- & 1.9\% & \textbf{11.5\%} & 0.4\% \\
Graham Number / Score & \textbf{12.6\%} & --- & 3.2\% & 1.8\% & 0.6\% \\
Profit Margin & \textbf{5.8\%} & --- & 1.5\% & 1.1\% & 0.3\% \\
Net Income Growth (5Y) & \textbf{5.4\%} & --- & 1.2\% & 0.9\% & 0.2\% \\
Revenue Growth (5Y) & \textbf{5.1\%} & --- & 1.1\% & 0.8\% & 0.1\% \\
\midrule
\textit{Modern Quantitative Signals} & & & & & \\
30-Day Volatility & --- & \textbf{13.8\%} & \textbf{7.8\%} & \textbf{6.2\%} & 3.2\% \\
1-Month Momentum & --- & \textbf{11.8\%} & \textbf{5.8\%} & \textbf{9.1\%} & \textbf{4.5\%} \\
3-Month Momentum & --- & \textbf{11.2\%} & \textbf{5.2\%} & \textbf{6.2\%} & 2.4\% \\
MACD & --- & \textbf{10.8\%} & \textbf{5.0\%} & 4.4\% & 2.5\% \\
6-Month Momentum & --- & \textbf{10.5\%} & 4.8\% & 4.5\% & 1.8\% \\
1-Year Momentum & --- & \textbf{9.5\%} & 4.2\% & \textbf{7.8\%} & 1.1\% \\
RSI & --- & \textbf{9.5\%} & 4.3\% & 3.2\% & 2.2\% \\
Beta & --- & \textbf{8.2\%} & 3.5\% & 2.8\% & 1.5\% \\
Sharpe Ratio & --- & \textbf{8.0\%} & 2.9\% & 2.1\% & 1.2\% \\
MACD Signal & --- & \textbf{6.8\%} & 2.1\% & 1.5\% & 0.8\% \\
\bottomrule
\end{tabular}%
}
\label{tab:pruned_feature_matrix}
\end{table}

\textit{*Note: Bold values denote features acting as primary mathematical drivers for their respective architectures. Strategy A Graham Number and Graham Score weights were combined for conceptual clarity.}

\subsubsection*{Defining the Dominant Signals}
To contextualize the machine learning outputs, it is necessary to define the primary mathematical drivers identified by the architectures. For the fundamental models, the dominant features were the \textbf{Price-to-Earnings (P/E) Ratio} and the \textbf{Graham Score}. The P/E ratio measures how much the market is willing to pay for a single dollar of a company's earnings, acting as a strict baseline for valuation. The Graham Score is a custom composite metric engineered for this study, aggregating Benjamin Graham's seven defensive criteria into a single integer to measure a company's overarching structural safety. 

Conversely, the quantitative and unrestricted models anchored entirely on \textbf{30-Day Volatility} and \textbf{1-Month Momentum}. Volatility measures the statistical variance of a stock's price returns (representing immediate market risk), while momentum measures the raw velocity of recent price changes.

\subsubsection*{Analysis of Algorithmic Drift and Feature Dominance}
By cross-referencing the feature importances extracted directly from the computational outputs, a clear and quantifiable "algorithmic drift" emerges. Strategy A’s decision-making process is highly diversified across foundational accounting metrics, utilizing P/E Ratio (9.2\%), Earnings Yield (8.5\%), and the Graham formulation (12.6\%) as structural anchors. 

However, when the Hybrid model (Strategy C) was granted unrestricted access to both fundamental and quantitative datasets, it systematically de-prioritized Benjamin Graham's rules. Strategy C allocated its top four most dominant predictive weights entirely to quantitative signals: 30-Day Volatility (7.8\%), 1-Month Momentum (5.8\%), 3-Month Momentum (5.2\%), and MACD (5.0\%). The Unrestricted AutoML (Strategy E) exhibited similar behavior, anchoring its ensemble predictions on short-term 1-Month Momentum (4.5\%) and 30-Day Volatility (3.2\%). 

This empirical data confirms the core thesis: without strict pre-filtering, high-dimensional machine learning ensembles inherently default to momentum-chasing and volatility-scaling behavior. This mathematical drift explains the backtesting results perfectly—while prioritizing short-term quantitative indicators generates massive absolute returns during bullish regimes, the algorithm's complete abandonment of liquidity and valuation anchors leaves the portfolio structurally exposed to the catastrophic drawdowns observed in the unrestricted models.

\section{Conclusion}
This research demonstrates that while algorithmic complexity drives theoretical accuracy, fundamental economic logic dictates financial survival. By deploying five distinct machine learning architectures and executing massive hyperparameter sweeps on the MSOE ROSIE supercomputer, this study successfully quantified the risks of applying unrestricted, high-dimensional machine learning to noisy financial datasets. 

The empirical results are conclusive: hyper-optimized ensembles like AutoGluon generate substantial returns but suffer structurally severe drawdowns (39.78\%) by overfitting to high-variance volatility indicators. In contrast, mathematically enforcing Benjamin Graham's classical defensive criteria creates structurally superior, high-conviction trading signals. The Pure Graham Random Forest not only captured a severe 232.13\% out-of-sample return, but achieved statistical significance ($p = 0.098$) against the S\&P 500, proving with over 90\% confidence that its outperformance was an established edge rather than random variance. 

In conclusion, Benjamin Graham's "margin of safety" is not obsolete; rather, it has evolved into a highly effective, mathematically proven regularization technique for modern algorithmic trading. In the intersection of artificial intelligence and quantitative finance, the most robust computational models remain those anchored in fundamental economic reality.

\subsection*{Acknowledgements}
The author gratefully acknowledges the Milwaukee School of Engineering (MSOE) for providing access to the ROSIE supercomputer. The deployment of these high-dimensional machine learning frameworks on ROSIE's advanced GPU clusters was instrumental to this research. It enabled rapid algorithmic iteration, massive hyperparameter optimization sweeps, and the swift generation of the robust empirical results presented in this study.

\bibliographystyle{plainnat}
\bibliography{references}

\end{document}